\documentclass{article}

\usepackage[nonatbib, final]{ewrl_2024}

\usepackage[utf8]{inputenc} 
\usepackage[T1]{fontenc}    
\usepackage{hyperref}       
\usepackage{url}            
\usepackage{booktabs}       
\usepackage{amsfonts}       
\usepackage{nicefrac}       
\usepackage{microtype}      
\usepackage{xcolor}         

\usepackage[numbers]{natbib}
\usepackage{amsmath}
\usepackage{amssymb}
\usepackage{mathtools}
\usepackage{amsthm}
\usepackage{pgfplots}
\pgfplotsset{compat=1.17}
\usepackage{graphicx}
\usepackage{enumitem}
\usepackage{caption}
\usepackage{float} 
\usepackage{afterpage}
\usepackage{placeins} 
\usepackage{microtype}
\usepackage{hyperref}
\usepackage{tikz}
\usetikzlibrary{positioning, arrows.meta}
\usetikzlibrary{decorations.markings}
\usetikzlibrary{shapes.geometric}

\title{Curricula for Learning Robust Policies with Factored State Representations in Changing Environments}

\author{%
  Panayiotis Panayiotou \\
  Department of Computer Science\\
  University of Bath\\
  Bath, United Kingdom \\
  \texttt{pp2024@bath.ac.uk} \\
  \And
  Özgür Şimşek \\
  Department of Computer Science\\
  University of Bath\\
  Bath, United Kingdom \\
  \texttt{o.simsek@bath.ac.uk} \\
}

\begin{document}

\maketitle

\begin{abstract}
Robust policies enable reinforcement learning agents to effectively adapt to and operate in unpredictable, dynamic, and ever-changing real-world environments. Factored representations, which break down complex state and action spaces into distinct components, can improve generalization and sample efficiency in policy learning. In this paper, we explore how the curriculum of an agent using a factored state representation affects the robustness of the learned policy. We experimentally demonstrate three simple curricula, such as varying only the variable of highest regret between episodes, that can significantly enhance policy robustness, offering practical insights for reinforcement learning in complex environments.

\end{abstract}

\section{Introduction}
\label{introduction}

Reinforcement learning has had remarkable success across a wide range of domains, including energy management \citep{wei2017deep}, robotic control \citep{stone2005reinforcement}, and strategic board games \citep{silver2016mastering}. However, in many applications, performance is evaluated solely on the training environment, often neglecting the importance of generalisation. This lack of emphasis contributes to some of the central challenges in reinforcement learning, including weak transferability between tasks and the brittleness of policies to small changes in environments or random seeds 
\citep{zhang2018study, farebrother2018generalization, lazaric2012transfer, taylor2009transfer}. Additionally, reinforcement learning algorithms often suffer from low sample efficiency, requiring large amounts of data to achieve robust performance.

Factored representations \citep{pearl1988probabilistic} decompose high-dimensional, unstructured state and action spaces into a few low-dimensional and high-level variables, each representing distinct and potentially independent aspects of the environment. This decomposition reduces the problem's dimensionality, possibly requiring fewer samples to learn a well-performing policy \citep{tsai2018learning, balaji2020factoredrl}. Additionally, factored representations can enhance a policy's ability to generalise across different parts of the state space, making it more robust and transferable \citep{baktashmotlagh2018learning, feng2022factored}.

\begin{figure}[h]
    \centering
    \includegraphics[width=0.65\linewidth]{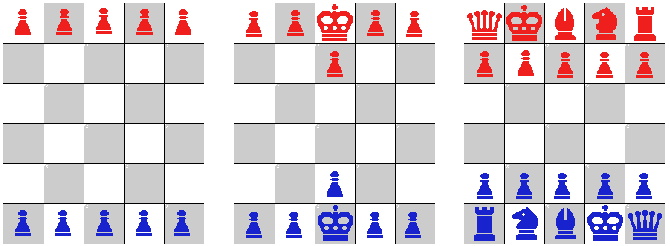}
    \caption{\fontsize{10pt}{11pt}\selectfont{\itshape{Quick Chess subgames, increasing in complexity from left to right (image source: \citet{narvekar2016source}).}}}
    \label{fig:quickchess}
\end{figure}

Curriculum learning \citep{bengio2009curriculum} is a training strategy that structures the learning process, such as by organising different subtasks in a particular sequence, with the goal of improving the learning speed or final performance. This can involve progressively increasing task difficulty or transferring knowledge between tasks of similar complexity. In reinforcement learning \citep{narvekar2017curriculum, narvekar2020curriculum, narvekar2016source}, this strategy involves training an agent on a sequence of different tasks, enabling it to leverage the knowledge gained from simpler tasks to tackle more challenging ones. This strategy can improve sample efficiency and enhance the robustness of the learned policies \citep{song2022robust}. For example, Quick Chess is a simplified version of chess that starts with easier subgames and gradually introduces the player to the whole game \citep{narvekar2016source}. As shown in Figure \ref{fig:quickchess}, early subgames can include only pawns to teach players how pawns move, attack and get promoted.

The real world is non-stationary and unpredictable, and we cannot capture all its variability in a static dataset or learning environment. In the real world, no two tasks are ever exactly the same, even if they may seem so in a simulation environment. Therefore, we aim to train robust policies that generalise effectively and adapt to unseen environments caused by distributional shifts \citep{kirk2023survey}. For example, a domain shift might change the position of an object between different runs, or a task shift might change the designated endpoint in a navigation task.

While factored representations can help in learning more robust policies \citep{baktashmotlagh2018learning, feng2022factored}, the role of curriculum learning in enhancing these policies remains underexplored. In this paper, we experimentally investigate how curriculum learning can improve the generalisation and adaptability of these policies to novel environments. We demonstrate the following:

\begin{enumerate}
    \item Without factored representations, simple curricula are insufficient for training robust policies that generalise well to unseen environments.
    \item Using factored representations, a curriculum of random shifts (domain randomisation) can enable learning robust policies.
    \item Using factored representations, a curriculum of shuffling a few diverse examples can allow learning robust policies.
    \item Using factored representations, we can design a curriculum for learning robust policies by identifying and adjusting the factors that cause the largest performance discrepancy (regret) when altered.
\end{enumerate}

\section{Preliminaries}
\label{preliminaries}

\paragraph{Markov Decision Processes.}
A Markov decision process (MDP) is a mathematical framework used to model decision-making problems. An MDP is defined by a tuple \((\mathcal{S}, \mathcal{A}, P, R, \gamma)\):
\begin{itemize}
    \item \(\mathcal{S}\) is a set of states.
    \item \(\mathcal{A}\) is a set of actions.
    \item \(P: \mathcal{S} \times \mathcal{A} \times \mathcal{S} \rightarrow [0,1]\) is a transition probability function, where \(P(s'|s,a)\) denotes the probability of transitioning to state \(s'\) from state \(s\) after taking action \(a\).
    \item \(R: \mathcal{S} \times \mathcal{A} \times \mathcal{S} \rightarrow \mathbb{R}\) is a reward function, where \(R(s,a, s')\) gives the expected reward for taking action \(a\) in state \(s\) and transitioning to state \(s'\).
    \item \(\gamma \in [0,1]\) is a discount factor.
\end{itemize}

\paragraph{Reinforcement Learning.}
Most commonly, the reinforcement learning problem is modelled as a Markov Decision Process. In this framework, a policy \( \pi(a|s) \) represents the probability of taking action \( a \) when the agent is in state \( s \). The objective is to learn a policy that maximises the expected cumulative return \( E_{\pi}[G_t] \), which is the sum of discounted rewards over time when following policy \(\pi\). The return \(G_t\) from time step \(t\) is defined as:
\[
G_t = \sum_{k=0}^{\infty} \gamma^k r_{t+k+1}.
\]
where $r_t$ indicates the reward at time step $t$. An agent interacts with its environment by taking actions based on its policy, receiving feedback in the form of rewards, and using this feedback to learn and improve its policy over time.

\paragraph{Dynamic Bayesian Networks.}
A Dynamic Bayesian Network is a probabilistic graphical model representing a set of variables and their conditional dependencies as a directed acyclic graph. It is specifically designed to model sequences of variables over time.
In a Dynamic Bayesian Network, the state at time \(t\), denoted \(\mathbf{X}_t = (X_{1,t}, X_{2,t}, \ldots, X_{n,t})\), depends on the state at time \(t-1\), denoted \(\mathbf{X}_{t-1}\). 
The joint probability distribution over \(\mathbf{X}_t\) is given by the product of the conditional probabilities of each variable given its parents in the graph, including temporal dependencies:
\[
P(\mathbf{X}_t \mid \mathbf{X}_{t-1}) = \prod_{i=1}^{n} P(X_{i,t} \mid \text{Pa}(X_{i,t})),
\]
where the parents of $X_{i,t}$, denoted as \(\text{Pa}(X_{i,t})\), include variables from both \(\mathbf{X}_{t-1}\) and \(\mathbf{X}_t\).

\paragraph{Factored Representations.}
Factored representations can decompose the state and action spaces into sets of variables, each representing different components of the environment. Formally, an atomic state \(s\) is represented as a vector of high-level factors \(s = (x_1, x_2, \ldots, x_n)\), and similarly an atomic action \(a = (y_1, y_2, \ldots, y_m)\).

\paragraph{Factorisation of MDPs.}
A Factored Markov decision process is a type of MDP in which the state space, action space, transition model, and sometimes the reward function are represented in a factored form. Factored models leverage structure in the problem to manage complexity. They can make solving larger MDPs more computationally feasible without losing accuracy because they represent the MDP more compactly, reducing the number of parameters. They can also generalise better in environments with large state or action spaces, allowing for more efficient policy learning and planning. 

The atomic state $s$ and atomic action $a$ can be represented as a factored representation of high-level factors. The transition probabilities \(P(s' | s, a)\) depend on a subset of state and action variables and are often represented using a Dynamic Bayesian Network. Similarly, the reward function can be defined as the sum of local reward functions $R_i$, depending only on a subset of the state and action variables. 

\paragraph{Distribution Shifts.}

Distribution shifts refer to changes in the data distribution encountered by an agent during different phases of learning, such as between training and testing. In reinforcement learning, the environment is often characterised by a set of variables that define its state and dynamics, such as the transition probabilities, reward functions, or physical properties (e.g. grid size in a grid-world task, friction coefficients in a robotic simulation, etc.). A distribution over environments refers to the probabilistic distribution of these variables. By sampling from this distribution, we obtain different instances of the environment, each with potentially different characteristics. Addressing distribution shifts is critical because real-world environments are typically non-stationary, meaning that the variables defining the environment can change over time.

We can distinguish between three different types of learning environments \citep{kirk2023survey}. First, there are singleton environments where the training and testing environments are identical. Secondly, there are independent and identically distributed (IID) environments where training and testing environments are different but from the same distribution. Thirdly, there are out-of-distribution environments where the training and testing environments are from different distributions.

\paragraph{Low-Regret Policies.} Regret is a measure of how much the performance of a policy (expected discounted cumulative reward) falls short of the optimal performance. Formally, the regret after \(T\) time steps of following policy \(\pi\) from an initial state $s_0$ can be defined as:
\[
Regret_{\pi}(s_0, T) = \sum_{t=0}^{T} \left( V^*(s_t) - V^{\pi}(s_t) \right),
\]

where $\pi^*$ is the optimal policy, $V^*(s_t)$ is the value function of the optimal policy at state $s_t$ representing the expected discounted cumulative reward from that state, and $V^\pi(s_t)$ is the value function of the current policy $\pi$ at state $s_t$.

We consider low-regret policies robust because a low regret ensures that the performance difference compared to the optimal policy is minimised, demonstrating the policy's ability to handle various scenarios and adapt to changes effectively.

\section{Background}
\label{background}

\begin{quote}
    ``No man ever steps in the same river twice.''
    \begin{flushright}
    -- Heraclitus
    \end{flushright}
\end{quote}

To effectively apply reinforcement learning in the real world, we must account for its non-stationary nature. Reflecting the idea of a constantly evolving environment, recent reinforcement learning research focuses on developing robust policies that can handle changing dynamics \citep{janner2019trust, gottesman2018evaluating, guo2022relational}, highlighting the need for policies that work in varied settings. 
Distribution shifts can significantly impact performance, leading to poor generalisation and arbitrarily high errors \citep{ross2012agnostic, richens2024robust}. For reinforcement learning to be successful in the real world, we must consider robustness and how shifts (e.g. an object changing colour) can impact both the domain \citep{farahani2021brief} and the task itself \citep{xian2018zero}.

Factored state representations, which involve breaking down the environment into distinct components, are an active area of research \citep{thomas2017independently, locatello2020weakly, bengio2013deep}. These representations have been shown to improve the sample efficiency of reinforcement learning algorithms in both tabular and deep reinforcement learning methods \citep{tsai2018learning, balaji2020factoredrl}. Additionally, they can help learn policies that are robust to domain shifts \citep{baktashmotlagh2018learning, feng2022factored}. It has also been proven that in scenarios where only the agent's decisions causally influence the reward (e.g. multi-armed bandits where the state does not affect the reward), all robust agents learn an approximate causal model \citep{richens2024robust}, which implies a factored representation.

Curriculum learning in reinforcement learning structures an agent's learning process by strategically ordering tasks that the agent experiences
\citep{narvekar2016source}.
It typically aims to enhance the agent's performance and learning speed by enabling the forward transfer of skills from simpler tasks to more challenging ones. 
A structured curriculum involves several key decisions: choosing the initial set of tasks, defining the progression of tasks, and establishing criteria for transitioning between them. Examples of such curricula include the work of \citet{silva2018object}, where tasks are randomly generated and grouped based on their ``transfer potential'', and \citet{narvekar2016source}, where a set of source tasks is continuously refined to match the agent's current abilities using methods like mistake-driven subtasks, which help the agent correct erroneous behaviour.
Similarly, unsupervised environment design \citep{dennis2020emergent} is a reinforcement learning training strategy that automatically generates a series of training environments to learn robust policies. Notable work in this area is ACCEL \citep{parker2022evolving}, which uses an evolutionary environment generator and regret-based feedback to make small edits to the environment and gradually introduce the agent to more complexity to train a robust policy.

\section{The Shifting Frozen Lake}
\label{task}

\begin{figure}[h]
    \centering
    \begin{minipage}{0.4\textwidth}
        \centering
        \includegraphics[width=0.8\linewidth]{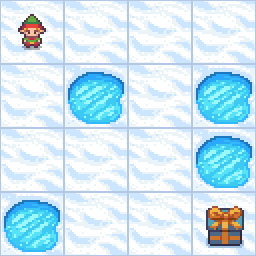}
    \end{minipage}\hfill
    \begin{minipage}{0.6\textwidth}
        
        \centering
        \begin{tikzpicture}[>=Stealth, node distance=0.5cm]
    
    \node[draw, ellipse, minimum width=3.8cm, minimum height=3.8cm, thick, align=center] (Ct) {
    hole\\locations\\
    $\begin{array}{cccc}
    0 & 0 & 0 & 0 \\
    0 & 1 & 0 & 1 \\
    0 & 0 & 0 & 1 \\
    1 & 0 & 0 & 0 \\
    \end{array}$
    };
    \node[draw, ellipse, right=of Ct, minimum width=1.2cm, minimum height=3.8cm, thick, align=center] (X4t) {
    distance\\
    matrix\\
    $\begin{array}{cccc}
    6 & 5 & 4 & 4 \\
    5 & \infty & 3 & \infty \\
    4 & 3 & 2 & \infty \\
    \infty & 2 & 1 & 0 \\
    \end{array}$
    };

    \node[draw, ellipse, below=of Ct, minimum width=1.2cm, minimum height=1.9cm, thick, align=center] (X2t) {
        grid size\\$4$
    };
    \node[draw, ellipse, right=of X2t, minimum width=1.2cm, minimum height=1.9cm, thick, align=center] (X3t) {agent\\location\\$(0, 0)$};
    \node[draw, ellipse, right=of X3t, minimum width=1.2cm, minimum height=0.7cm, thick, align=center] (X1t) {goal\\location\\$(3, 3)$};
    \end{tikzpicture}
    \end{minipage}
    \caption{\fontsize{10pt}{11pt}\selectfont{\itshape{A sample Frozen Lake environment. On the right hand side, we present a factored representation of the state. Using this factored representation, the transition function can be factorised using a Dynamic Bayesian Network (see Figure \ref{fig:sfl-dbn} in Appendix \ref{appendix:a}).
    }}}
    \label{fig:frozen-lake-preview}
\end{figure}

We define the Shifting Frozen Lake environment, where aspects of the environment can exhibit a shifting behaviour, allowing us to test for out-of-distribution generalisation. 

Frozen Lake \citep{1606.01540} is a grid-world environment where the agent navigates from a designated start cell (top-left) to a goal cell (bottom-right). The agent can move up, down, left, or right, and must avoid falling into holes along the way. Depending on the configuration, the actions can be either stochastic or deterministic. For an example, refer to Figure \ref{fig:frozen-lake-preview}.

In the original Frozen Lake environment, the start location, goal location, hole locations, and grid size are kept constant throughout all the episodes. In Shifting Frozen Lake, the grid size N, the positions of the holes, the starting point, and the goal location can change from one episode to the next. For simplicity, we assume that these variables remain constant during an episode despite potential changes, such as warm weather that could cause the lake to start melting. Due to the changing nature of the environment (e.g. the start location might change), we will refer to different instances of the environment as ``examples''.

Below is a full specification of the task:

\begin{itemize}
\item Actions: Left, down, right, up with deterministic transitions.

\item State: This is an $N \times N$ matrix that shows where the agent, the goal, the holes and the frozen squares are. We can factorise this using the 5-tuple (grid size, hole locations, agent location, goal location, distance matrix) as seen in Figure \ref{fig:frozen-lake-preview}.

\item Start state: An initial location $[x_I, x_I]$ where $N\times N$ are the dimensions of the grid, and $0 \leq x_I, y_I < N$.

\item Goal state: A location $[x_G, y_G]$ where $N\times N$ are the dimensions of the grid, and $0 \leq x_G, y_G < N$. All examples always include a possible path from the start state to the goal state.

\item Rewards: $-0.1$ for each move, an additional $+10$ for reaching the goal, and an additional $-10$ for reaching a hole. The discount factor is \( \gamma = 1 \). 

\item The episode ends if the player moves into a hole or the goal state.

\end{itemize}

\paragraph{Environment shifts.} The initial location, goal location, hole locations, and grid size can change from episode to episode. The environment supports the following shifting behaviours and the functionality to switch between them:
        \begin{itemize}
            \item \textbf{No Shifting}: The variables are sampled once upon the creation of the environment and remain constant for all episodes. 
            \item \textbf{Random Shifting}: At the start of each new episode, the environment uniformly resamples all variables (start location, goal location, hole locations, grid size).
            \item \textbf{Single Preset Variable Shifting}: One variable is specified to shift. Upon the creation of the environment, all variables are sampled once. In each episode, only the chosen variable is resampled.
            \item \textbf{Single Random Variable Shifting}: Upon creation, all variables are sampled. In each episode, one randomly chosen variable is resampled, changed, and reverted at the end of the episode.
            \item \textbf{Stored Examples Shifting}: Upon creation, a sample of \(N\) examples is stored. For each new episode, one of these examples is randomly selected and used.
    \end{itemize}

The state can be factorised by using variables that denote the grid size, hole locations, goal location, the current agent location, and a distance matrix from the goal location. However, this factorisation has redundancies, e.g. the hole locations can be inferred from the distance matrix. We can optimise the factored representation by retaining only the relevant variables, reducing redundancy and improving efficiency. For example, using only the distance matrix and the current agent location, an agent can learn an optimal and robust policy by always taking the shortest path to the end.

\section{Experiments}
\label{experiments}

Our experiments include the following agents:

\begin{itemize}
    \item \textbf{Random Action Selection}: Selects action uniformly at random. Used as a baseline.
    \item \textbf{Optimal}: Achieves the highest possible performance by using breadth-first search to pick the direction with the smallest distance to the goal (without falling into a hole).
    \item \textbf{PPO}: Without using a factored representation, we apply a convolutional neural network to the grid, where each tile is one-hot encoded in a separate channel. We pad the grid with a special character so all grids have the same size.
    \item \textbf{PPO-F}: A PPO agent using an optimised factored representation, retaining only the immediate neighbourhood in the distance matrix, which is sufficient for the agent to act optimally in this task. The agent does not model the transition function or use the assumption that the transition function can be factorised.
\end{itemize}

We run all the experiments for five agents and plot the mean and standard error of the total undiscounted reward per epoch ($\gamma=1$). Each epoch consists of $900$ time steps, and each episode has a timeout of $100$ time steps. 
Performance scores around $-30$ indicate ``stuck'' behaviour, where agents avoid losses by engaging in repetitive, looping movements, such as endlessly alternating between left and right actions. Scores higher than $-30$ but worse than optimal performance indicate an agent that solves some of the grids. For these experiments, we consider random shifting (resampling all variables at the start of each episode) as a test of deep understanding and generalisation of the task because it requires agents to know how to navigate to the goal from anywhere and avoid holes.

We explore the following curricula, with changes in curriculum phases indicated by vertical dotted lines in the figures:
\begin{enumerate}[label=(\Alph*)]
    \item \textbf{No Shifting to Random Shifting}: Fit a single example, then shift all variables randomly to test generalisation.
    
    \item \textbf{No Shifting to Single Random Variable Shifting}: Fit a single example and then randomly shift only one variable per episode.

    \item \textbf{Random Shifting}: Test generalisation from diverse training (domain randomisation) by shifting all variables randomly from the start.
    
    \item \textbf{Stored Examples to Random Shifting}: Train a policy by shuffling a few pre-sampled examples and then test generalisation by shifting all variables randomly.
    
    \item \textbf{Single Preset Variable Shifting to Random Shifting}: Shift only one specified variable initially, then shift all variables randomly to test generalisation. 

\end{enumerate}

\newlength{\myfigurewidth}
\setlength{\myfigurewidth}{0.48\linewidth}

\FloatBarrier
\afterpage{
\clearpage

\begin{figure}[H]
    \centering
    \begin{minipage}{\myfigurewidth}
        \centering
        \includegraphics[trim=0.4cm 0cm 3.5cm 2.4cm, clip, width=\linewidth]{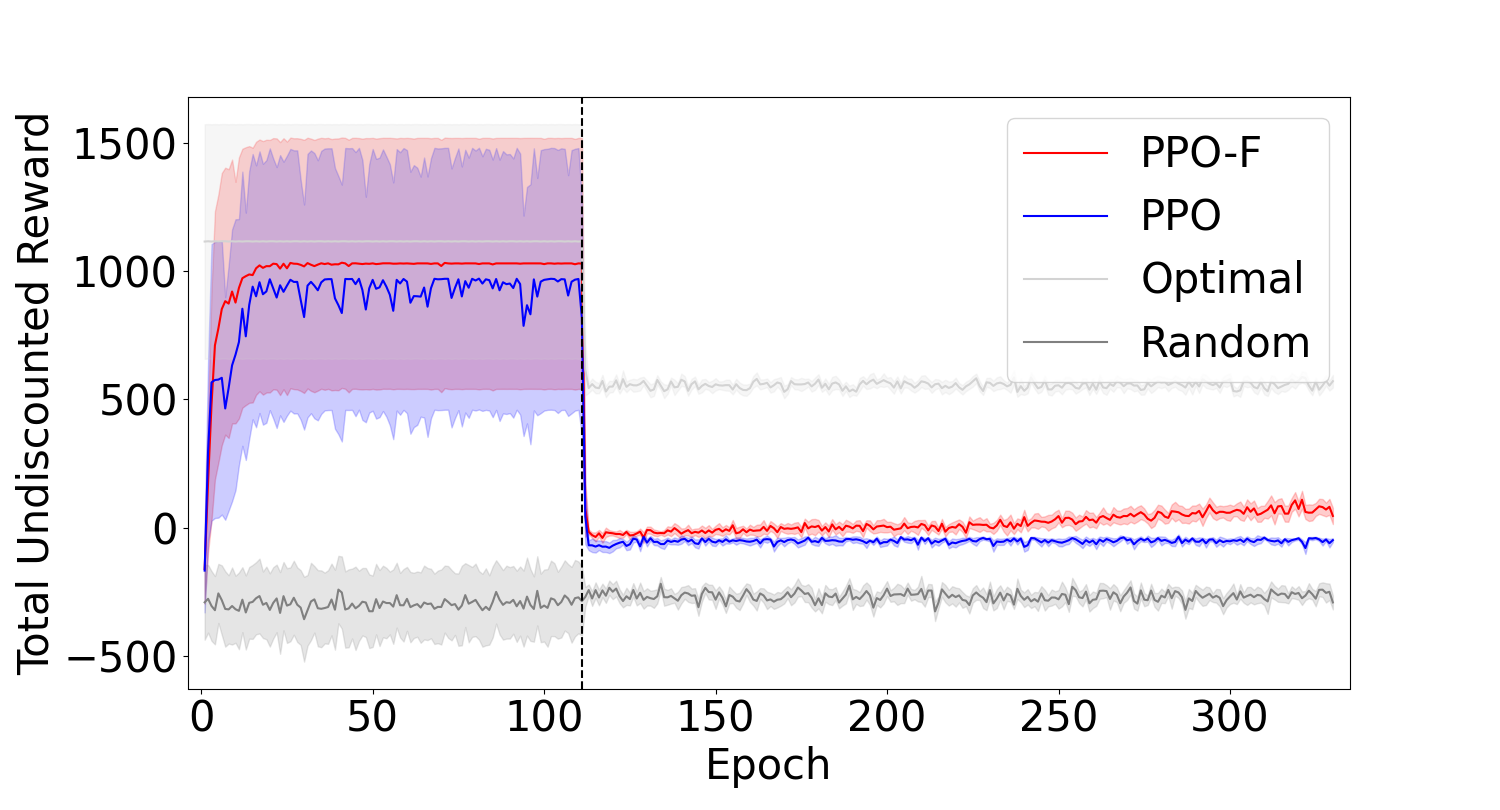}
        \caption{\fontsize{10pt}{11pt}\selectfont{\itshape{Curriculum (A): No Shifting to Random Shifting.}}}        
        \label{fig:exp3}
    \end{minipage}\hfill 
    \begin{minipage}{\myfigurewidth}
        \centering
        \includegraphics[trim=0.4cm 0cm 3.5cm 2.4cm, clip, width=\linewidth]{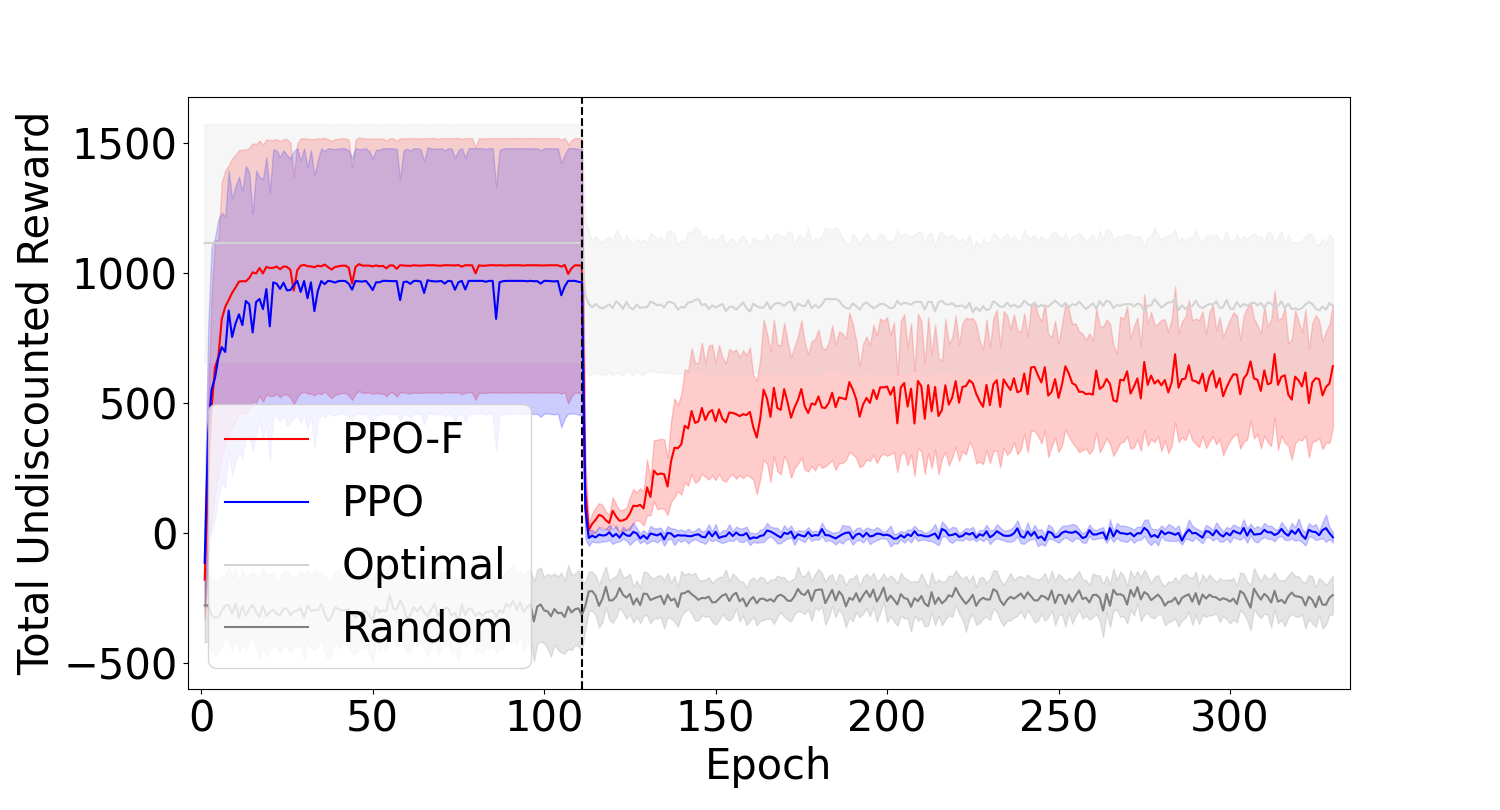}
        \caption{\fontsize{10pt}{11pt}\selectfont{\itshape{Curriculum (B): No Shifting to Single Random Variable Shifting.}}}
        \label{fig:exp4}
    \end{minipage}
\end{figure}

\begin{figure}[H]
    \centering
    \begin{minipage}{\myfigurewidth}
        \centering
        \includegraphics[trim=0.4cm 0cm 3.5cm 2.4cm, clip, width=\linewidth]{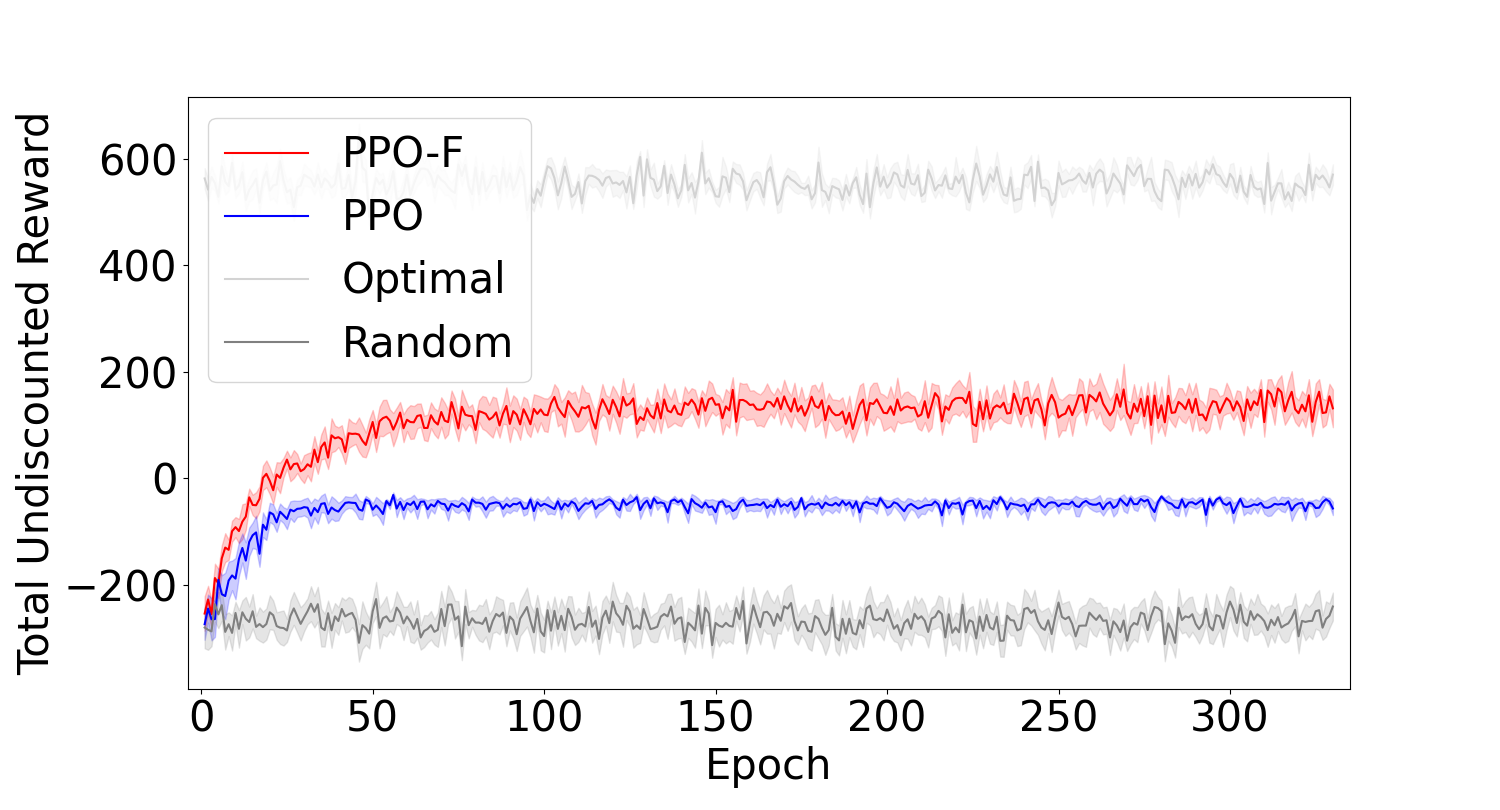}
        \caption{\fontsize{10pt}{11pt}\selectfont{\itshape{Curriculum (C): Random Shifting.}}}
        \label{fig:exp8}
    \end{minipage}\hfill 
    \begin{minipage}{\myfigurewidth}
        \centering
        \includegraphics[trim=0.4cm 0cm 3.5cm 2.4cm, clip, width=\linewidth]
        {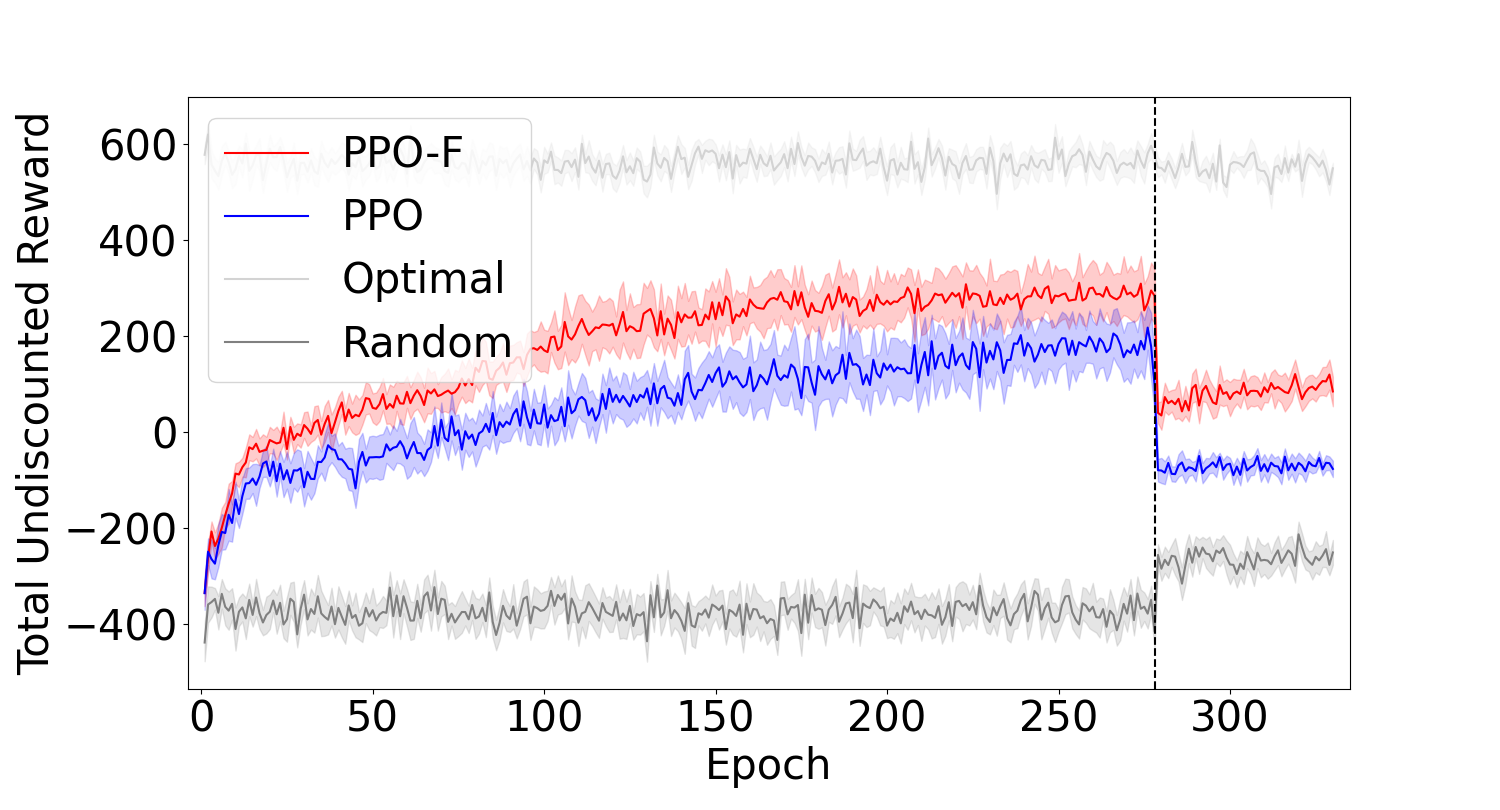}
        \caption{\fontsize{10pt}{11pt}\selectfont{\itshape{Curriculum (D): 15 Stored Examples to Random Shifting.}}}
        \label{fig:exp5}
    \end{minipage}
\end{figure}

\begin{figure}[H]
    \centering
    \begin{minipage}{\myfigurewidth}
        \centering

        \includegraphics[trim=0.4cm 0cm 3.5cm 2.4cm, clip, width=\linewidth]
        {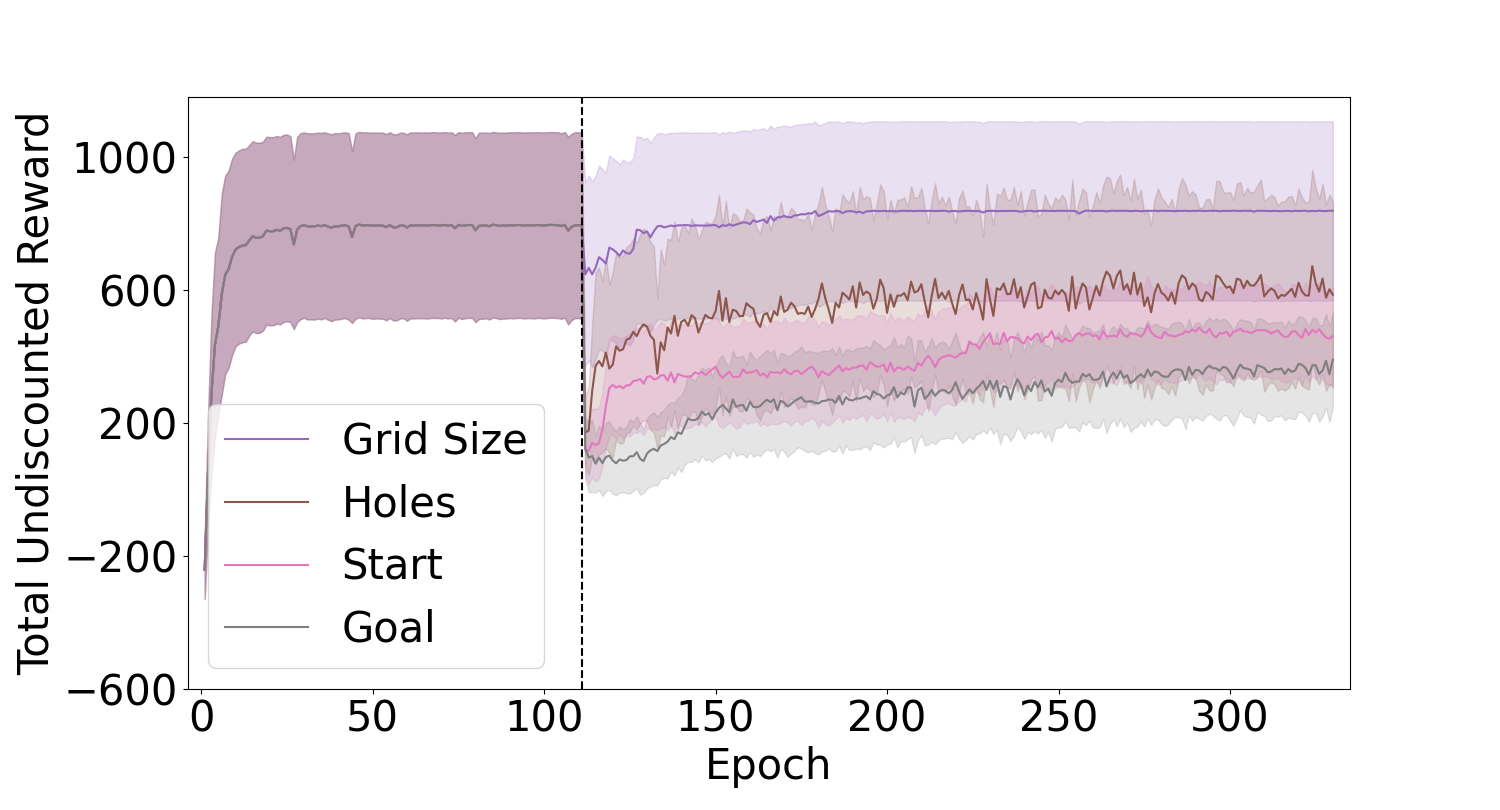}
        \caption{\fontsize{10pt}{11pt}\selectfont{\itshape{Preliminary Experiment for Curriculum (E): Fit PPO-F to a single example, then shift only one of the four variables.}}}
        \label{fig:one_train}
        
    \end{minipage}\hfill 
    \begin{minipage}{\myfigurewidth}
        \centering
        
        \includegraphics[trim=0.4cm 0cm 3.5cm 2.4cm, clip, width=\linewidth]{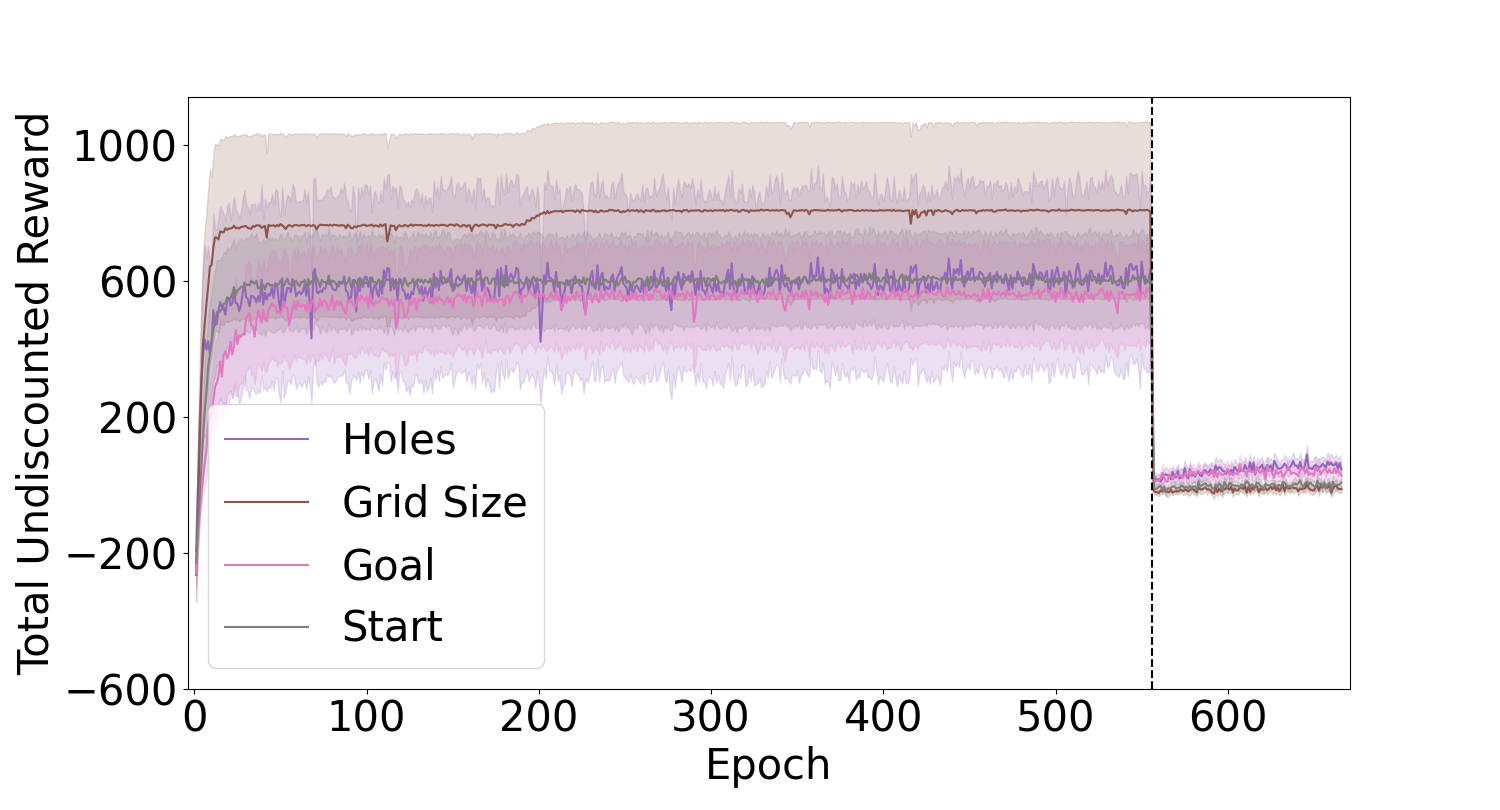}
        \caption{\fontsize{10pt}{11pt}\selectfont{\itshape{Curriculum (E): Single Preset Variable Shifting to Random Shifting on PPO-F.}}}
        \label{fig:onetrain}
        
    \end{minipage}
\end{figure}



\begin{figure}[H]
    \centering
    \begin{minipage}{\myfigurewidth}
        \centering

        \includegraphics[trim=0cm 0cm 0cm 0.5cm, clip, width=0.9\linewidth]{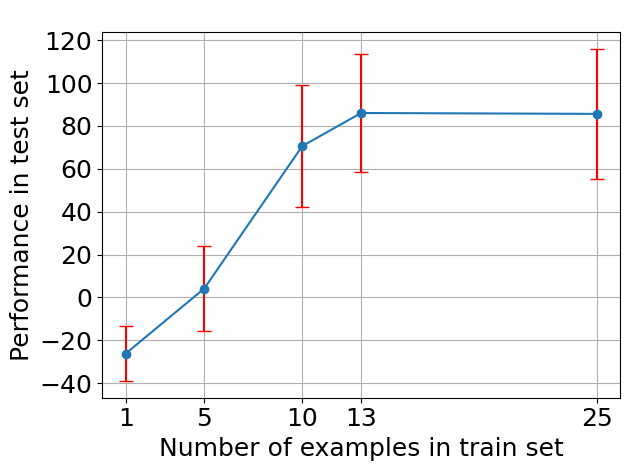}
        \caption{\fontsize{10pt}{11pt}\selectfont{\itshape{
        PPO-F performance after training with different numbers of stored examples, followed by 50 epochs of random shifting (test).
        }}}
        \label{fig:vs_plot}

    \end{minipage}\hfill 
    \begin{minipage}{\myfigurewidth}
        \centering
        
        \includegraphics[trim=0.4cm 0cm 3.5cm 2cm, clip, width=\linewidth]{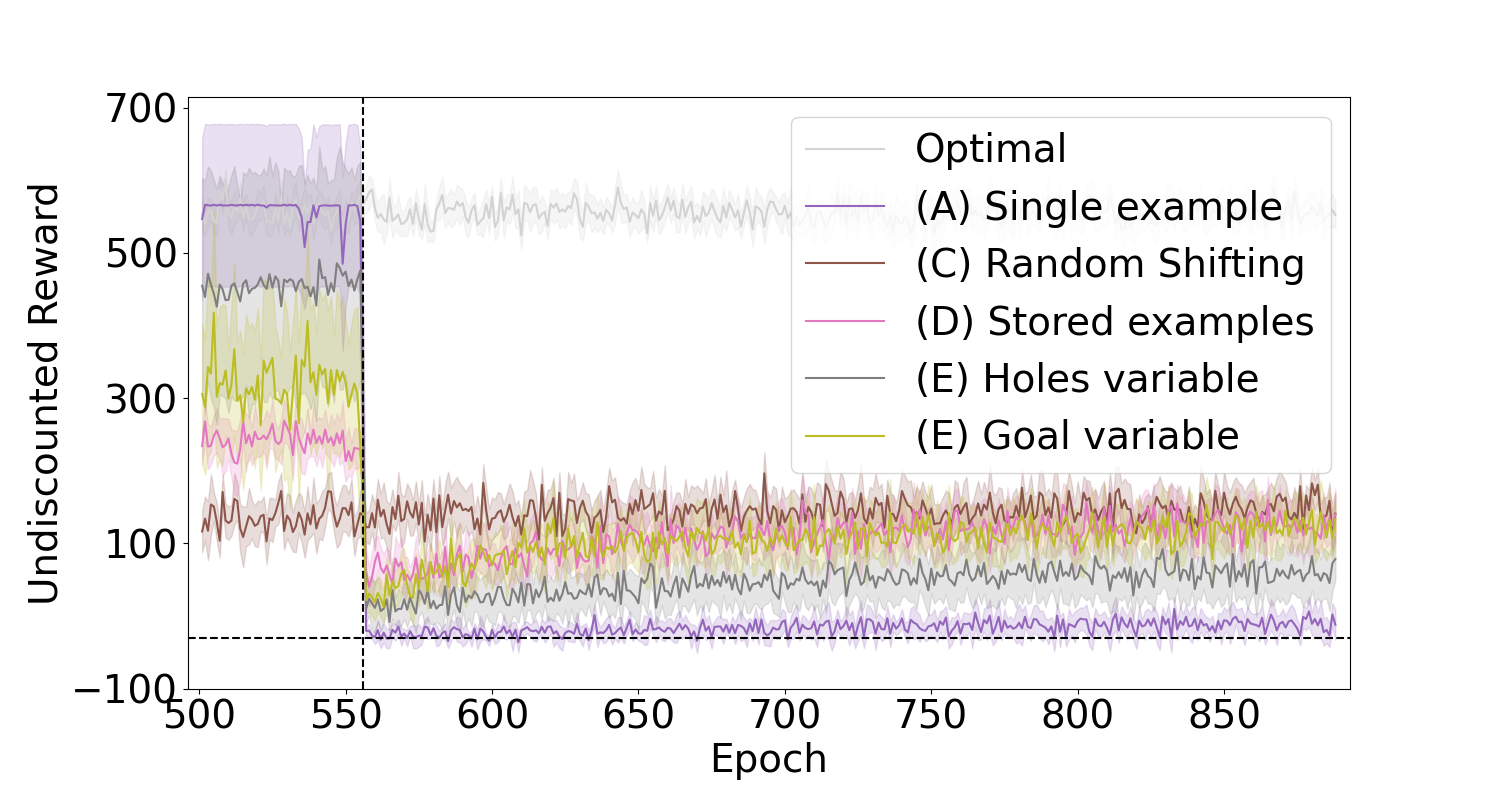}
        \caption{\fontsize{10pt}{11pt}\selectfont{\itshape{Comparing Curricula for Factored Agents.}}}
        \label{fig:comparethemall}
        
    \end{minipage}
\end{figure}

        \begin{figure}[H]
            \centering
            \begin{tikzpicture}
                \begin{axis}[
                    ybar,
                    bar width=20pt,
                    ylabel={Regret},
                    enlarge x limits=0.25,
                    ymin=0,
                    ylabel near ticks,
                    symbolic x coords={C, D, E-holes, E-goal},
                    xtick=data,
                    xticklabels={C, D, E-holes, E-goal},
                    ymajorgrids=true,
                    grid style=dashed,
                ]
                    \addplot+ [
                        error bars/.cd,
                        y dir=both,
                        y explicit,
                    ] coordinates {
                        (C,4605.92) +- (0,80.25)
                        (D,5769.30) +- (0,290.86)
                        (E-holes,6745.08) +- (0,571.39)
                        (E-goal,6942.48) +- (0,891.01)
                    };
                \end{axis}
            \end{tikzpicture}
            \caption{\fontsize{10pt}{11pt}\selectfont{\itshape{Comparing the regret of robust policies learnt by following Curricula C, D, E with varying holes, and E with a varying goal location.}}}
            \label{fig:regret}
        \end{figure}

}
\FloatBarrier

\paragraph{Curriculum (A): No Shifting to Random Shifting.} We test generalisation from a single example and present the results in Figure \ref{fig:exp3}. When fitting a single example, the methods show a significant standard error because the grid size can vastly change the reward per epoch. For instance, reaching the goal in 3 steps on a 4x4 grid gives 970 points per epoch, while 20 steps on a 10x10 grid give only 120 points per epoch. None of the trained methods demonstrate significant knowledge transfer after the shift, as their performance drops to around 0. After the shift, \textit{PPO} exhibits ``stuck'' behaviour, repeatedly moving left/right or up/down. \textit{PPO-F} is more active but only solves about 25\% of the examples right after the shift. It loses in around 9\% of the examples and displays ``stuck'' behaviour in the rest.

\paragraph{Curriculum (B): No Shifting to Single Random Variable Shifting.} In Figure \ref{fig:exp4}, we see that both \textit{PPO} and \textit{PPO-F} exhibit similarly low knowledge transfer and robustness as there is a big performance drop when random shifting starts. Notably, after a few epochs with random shifting, \textit{PPO-F} adapts quickly to the new task distribution.

\paragraph{Curriculum (C): Random Shifting.} In Figure \ref{fig:exp8}, we evaluate how well the agents can generalise from diverse training. The test and train distributions of the environment here are identical, so this is IID generalisation. Note, however, that diverse training complicates the learning task. A closer examination of the \textit{PPO} agent behaviour reveals that it often fails to reach the goal in any episode, reverting to ``stuck'' behaviour. Factored variables, however, provide a significant advantage in discovering the task structure. The \textit{PPO-F} agent identifies a robust policy within 75 epochs. 
But its policy is far from optimal. We examined the agent's performance over 50 epochs following stabilisation and found that 19\% of its movements were repetitive, back-and-forth motions. It executed $3.6\times$ more moves than the optimal agent and, on average, fell into a hole 24.8 times per epoch.

\paragraph{Curriculum (D): Stored Examples to Random Shifting.} In Figure \ref{fig:exp5}, we evaluate how well the agent generalises using only 15 training examples. 
\textit{PPO} exhibits ``stuck'' behaviour after the shift and shows no signs of knowledge transfer or robustness. \textit{PPO-F} shows strong knowledge transfer and robustness, performing on par with diverse training after seeing only 15 examples. We examined the agent's performance over 50 epochs following stabilisation and found that 35\% of its movements were repetitive, back-and-forth motions. It executed $4.5\times$ more moves than the optimal agent and, on average, fell into a hole 7.8 times per epoch. This experiment demonstrates that a few diverse examples are sufficient to build a robust policy over a factored state representation.

\paragraph{Curriculum (D) Follow-Up.} We further investigate in Figure \ref{fig:vs_plot} how many stored examples are needed to achieve good test performance under Random Shifting. We train multiple \textit{PPO-F} agents with different numbers of training examples and estimate their test performance by averaging over 50 epochs under Random Shifting. Generally, we expect more diverse training with more examples to correlate with improved performance. However, there are diminishing returns, as fitting more examples takes longer and does not necessarily result in better performance. Performance gains level off after fitting 13 examples, and training with more examples significantly increases the training time.

\paragraph{Preliminary Experiment for Curriculum (E).} We train \textit{PPO-F} on a single example and then shift only a specific variable on each episode (Single Preset Variable Shifting), as shown in Figure \ref{fig:one_train}. We find that shifting the goal location, start location, and hole locations leads to high regret while shifting the grid size does not. In Curriculum (E), we then investigate if shifting only one of these during training is enough to learn a robust policy.

\paragraph{Curriculum (E): Single Preset Variable Shifting to Random Shifting.} In Figure \ref{fig:onetrain}, we evaluate how well the \textit{PPO-F} agent generalises when only one variable is shifted during training. We examine four training curricula, each shifting only one variable (holes, grid size, goal location, or start location). We test OOD generalisation by exposing the agents to random shifts. We find that varying just one variable, either hole locations or goal location, leads to learning a robust policy. Two of the three variables that cause high regret are sufficient by themselves when shifted to form a curriculum for training a robust agent.
We examined the agents' performance over 50 epochs following stabilisation and found that when changing the holes, they executed $6.8\times$ more moves than the optimal agent and, on average, fell into a hole 5.2 times per epoch. When changing the goal location, they executed $8.8\times$ more moves than the optimal agent and, on average, fell into a hole 5.4 times per epoch. 

\paragraph{Comparing Curricula for Factored Agents.} In Figure \ref{fig:comparethemall}, we compare \textit{PPO-F} agents trained with four robust curricula (C, D, E with varying holes, and E with varying goals) and a single-example curriculum (A). All robust curricula outperform the single-example curriculum after the start of random shifting.
Training with random shifts gives the best immediate post-shift performance. However, the other robust curricula perform similarly. Curricula (D) and (E) (with a varying goal) adapt quickly and reach the same performance as (C). The horizontal dotted line represents the ``stuck'' behaviour observed by the single-example curriculum after the shift.
Pre-shift performance is not comparable between agents because each curriculum exposes agents to different environments, and smaller grids lead to higher total rewards per epoch.

\paragraph{Regret Analysis of Robust Policies.} In Figure \ref{fig:regret}, we rank the four robust curricula (C, D, E-holes, E-goals) by increasing regret and decreasing robustness (from left to right). This ranking also reflects decreasing risk-taking and falling in holes. Curriculum (C) benefits from test and train environments being identical (IID generalisation). The two (E) curricula only modify one variable at a time but learn robust policies. 
We suggest that changing the variable that shifts during training could further enhance policy robustness.

\section{Discussion}
\label{discussion}

First, our results demonstrate that methods using factored representations can help learn robust policies more easily. Agents using an atomic state representation usually fail to reach the goal when the environment has distribution shifts. While a tailored curriculum could help such agents to learn robust policies, simple curricula may be enough for agents that use a factored representation. 
Secondly, the curriculum used significantly impacted the robustness of the learned policy over a factored state representation. 
The agents learned comparably robust policies with either diverse training, shuffling a few stored examples, or by shifting a single variable that caused high regret when altered alone (true for two out of three variables). We also quantitatively compare the robustness of the learned policies following each of the curricula and point out the effect of the curriculum on the risk aversion and performance of the learned policies. 
Lastly, we believe that enabling agents to autonomously generate their own curricula by identifying and adjusting variables that require further exploration (such as those causing high regret) will lead to learning even more robust policies and better generalization across diverse environments.

\section*{Acknowledgements}

This work was supported by the UKRI Centre for Doctoral Training in Accountable, Responsible and Transparent AI (ART-AI) [EP/S023437/1]. We thank Joshua B. Evans for useful discussions.




\bibliographystyle{plainnat}
\bibliography{refs}

\begin{thebibliography}{32}
\providecommand{\natexlab}[1]{#1}
\providecommand{\url}[1]{\texttt{#1}}
\expandafter\ifx\csname urlstyle\endcsname\relax
  \providecommand{\doi}[1]{doi: #1}\else
  \providecommand{\doi}{doi: \begingroup \urlstyle{rm}\Url}\fi

\bibitem[Baktashmotlagh et~al.(2018)Baktashmotlagh, Faraki, Drummond, and Salzmann]{baktashmotlagh2018learning}
Mahsa Baktashmotlagh, Masoud Faraki, Tom Drummond, and Mathieu Salzmann.
\newblock Learning factorized representations for open-set domain adaptation.
\newblock \emph{arXiv preprint arXiv:1805.12277}, 2018.

\bibitem[Balaji et~al.(2020)Balaji, Christodoulou, lu, Jeon, and Bell-Masterson]{balaji2020factoredrl}
Bharathan Balaji, Petros Christodoulou, Xiaoyu lu, Byungsoo Jeon, and Jordan Bell-Masterson.
\newblock Factoredrl: Leveraging factored graphs for deep reinforcement learning.
\newblock In \emph{NeurIPS 2020 Workshop on Deep Reinforcement Learning}, 2020.

\bibitem[Bengio(2013)]{bengio2013deep}
Yoshua Bengio.
\newblock Deep learning of representations: Looking forward.
\newblock In \emph{International conference on statistical language and speech processing}, pages 1--37. Springer, 2013.

\bibitem[Bengio et~al.(2009)Bengio, Louradour, Collobert, and Weston]{bengio2009curriculum}
Yoshua Bengio, J{\'e}r{\^o}me Louradour, Ronan Collobert, and Jason Weston.
\newblock Curriculum learning.
\newblock In \emph{Proceedings of the 26th annual international conference on machine learning}, pages 41--48, 2009.

\bibitem[Brockman et~al.(2016)Brockman, Cheung, Pettersson, Schneider, Schulman, Tang, and Zaremba]{1606.01540}
Greg Brockman, Vicki Cheung, Ludwig Pettersson, Jonas Schneider, John Schulman, Jie Tang, and Wojciech Zaremba.
\newblock Openai gym, 2016.

\bibitem[Dennis et~al.(2020)Dennis, Jaques, Vinitsky, Bayen, Russell, Critch, and Levine]{dennis2020emergent}
Michael Dennis, Natasha Jaques, Eugene Vinitsky, Alexandre Bayen, Stuart Russell, Andrew Critch, and Sergey Levine.
\newblock Emergent complexity and zero-shot transfer via unsupervised environment design.
\newblock \emph{Advances in neural information processing systems}, 33:\penalty0 13049--13061, 2020.

\bibitem[Farahani et~al.(2021)Farahani, Voghoei, Rasheed, and Arabnia]{farahani2021brief}
Abolfazl Farahani, Sahar Voghoei, Khaled Rasheed, and Hamid~R Arabnia.
\newblock A brief review of domain adaptation.
\newblock \emph{Advances in data science and information engineering: proceedings from ICDATA 2020 and IKE 2020}, pages 877--894, 2021.

\bibitem[Farebrother et~al.(2018)Farebrother, Machado, and Bowling]{farebrother2018generalization}
Jesse Farebrother, Marlos~C Machado, and Michael Bowling.
\newblock Generalization and regularization in dqn.
\newblock \emph{arXiv preprint arXiv:1810.00123}, 2018.

\bibitem[Feng et~al.(2022)Feng, Huang, Zhang, and Magliacane]{feng2022factored}
Fan Feng, Biwei Huang, Kun Zhang, and Sara Magliacane.
\newblock Factored adaptation for non-stationary reinforcement learning.
\newblock \emph{Advances in Neural Information Processing Systems}, 35:\penalty0 31957--31971, 2022.

\bibitem[Gottesman et~al.(2018)Gottesman, Johansson, Meier, Dent, Lee, Srinivasan, Zhang, Ding, Wihl, Peng, et~al.]{gottesman2018evaluating}
Omer Gottesman, Fredrik Johansson, Joshua Meier, Jack Dent, Donghun Lee, Srivatsan Srinivasan, Linying Zhang, Yi~Ding, David Wihl, Xuefeng Peng, et~al.
\newblock Evaluating reinforcement learning algorithms in observational health settings.
\newblock \emph{arXiv preprint arXiv:1805.12298}, 2018.

\bibitem[Guo et~al.(2022)Guo, Gong, and Tao]{guo2022relational}
Jixian Guo, Mingming Gong, and Dacheng Tao.
\newblock A relational intervention approach for unsupervised dynamics generalization in model-based reinforcement learning.
\newblock \emph{arXiv preprint arXiv:2206.04551}, 2022.

\bibitem[Janner et~al.(2019)Janner, Fu, Zhang, and Levine]{janner2019trust}
Michael Janner, Justin Fu, Marvin Zhang, and Sergey Levine.
\newblock When to trust your model: Model-based policy optimization.
\newblock \emph{Advances in neural information processing systems}, 32, 2019.

\bibitem[Kirk et~al.(2023)Kirk, Zhang, Grefenstette, and Rockt{\"a}schel]{kirk2023survey}
Robert Kirk, Amy Zhang, Edward Grefenstette, and Tim Rockt{\"a}schel.
\newblock A survey of zero-shot generalisation in deep reinforcement learning.
\newblock \emph{Journal of Artificial Intelligence Research}, 76:\penalty0 201--264, 2023.

\bibitem[Lazaric(2012)]{lazaric2012transfer}
Alessandro Lazaric.
\newblock Transfer in reinforcement learning: a framework and a survey.
\newblock In \emph{Reinforcement Learning: State-of-the-Art}, pages 143--173. Springer, 2012.

\bibitem[Locatello et~al.(2020)Locatello, Poole, R{\"a}tsch, Sch{\"o}lkopf, Bachem, and Tschannen]{locatello2020weakly}
Francesco Locatello, Ben Poole, Gunnar R{\"a}tsch, Bernhard Sch{\"o}lkopf, Olivier Bachem, and Michael Tschannen.
\newblock Weakly-supervised disentanglement without compromises.
\newblock In \emph{International conference on machine learning}, pages 6348--6359. PMLR, 2020.

\bibitem[Narvekar(2017)]{narvekar2017curriculum}
Sanmit Narvekar.
\newblock Curriculum learning in reinforcement learning.
\newblock In \emph{IJCAI}, pages 5195--5196, 2017.

\bibitem[Narvekar et~al.(2016)Narvekar, Sinapov, Leonetti, and Stone]{narvekar2016source}
Sanmit Narvekar, Jivko Sinapov, Matteo Leonetti, and Peter Stone.
\newblock Source task creation for curriculum learning.
\newblock In \emph{Proceedings of the 2016 international conference on autonomous agents \& multiagent systems}, pages 566--574, 2016.

\bibitem[Narvekar et~al.(2020)Narvekar, Peng, Leonetti, Sinapov, Taylor, and Stone]{narvekar2020curriculum}
Sanmit Narvekar, Bei Peng, Matteo Leonetti, Jivko Sinapov, Matthew~E Taylor, and Peter Stone.
\newblock Curriculum learning for reinforcement learning domains: A framework and survey.
\newblock \emph{Journal of Machine Learning Research}, 21\penalty0 (181):\penalty0 1--50, 2020.

\bibitem[Parker-Holder et~al.(2022)Parker-Holder, Jiang, Dennis, Samvelyan, Foerster, Grefenstette, and Rockt{\"a}schel]{parker2022evolving}
Jack Parker-Holder, Minqi Jiang, Michael Dennis, Mikayel Samvelyan, Jakob Foerster, Edward Grefenstette, and Tim Rockt{\"a}schel.
\newblock Evolving curricula with regret-based environment design.
\newblock In \emph{International Conference on Machine Learning}, pages 17473--17498. PMLR, 2022.

\bibitem[Pearl(1988)]{pearl1988probabilistic}
Judea Pearl.
\newblock \emph{Probabilistic reasoning in intelligent systems: networks of plausible inference}.
\newblock Morgan kaufmann, 1988.

\bibitem[Richens and Everitt(2024)]{richens2024robust}
Jonathan Richens and Tom Everitt.
\newblock Robust agents learn causal world models.
\newblock \emph{arXiv preprint arXiv:2402.10877}, 2024.

\bibitem[Ross and Bagnell(2012)]{ross2012agnostic}
Stephane Ross and J~Andrew Bagnell.
\newblock Agnostic system identification for model-based reinforcement learning.
\newblock \emph{arXiv preprint arXiv:1203.1007}, 2012.

\bibitem[Silva and Costa(2018)]{silva2018object}
Felipe Leno~Da Silva and Anna Helena~Reali Costa.
\newblock Object-oriented curriculum generation for reinforcement learning.
\newblock In \emph{Proceedings of the 17th international conference on autonomous agents and multiagent systems}, pages 1026--1034, 2018.

\bibitem[Silver et~al.(2016)Silver, Huang, Maddison, Guez, Sifre, Van Den~Driessche, Schrittwieser, Antonoglou, Panneershelvam, Lanctot, et~al.]{silver2016mastering}
David Silver, Aja Huang, Chris~J Maddison, Arthur Guez, Laurent Sifre, George Van Den~Driessche, Julian Schrittwieser, Ioannis Antonoglou, Veda Panneershelvam, Marc Lanctot, et~al.
\newblock Mastering the game of go with deep neural networks and tree search.
\newblock \emph{nature}, 529\penalty0 (7587):\penalty0 484--489, 2016.

\bibitem[Song and Schneider(2022)]{song2022robust}
Yeeho Song and Jeff Schneider.
\newblock Robust reinforcement learning via genetic curriculum.
\newblock In \emph{2022 International Conference on Robotics and Automation (ICRA)}, pages 5560--5566. IEEE, 2022.

\bibitem[Stone et~al.(2005)Stone, Sutton, and Kuhlmann]{stone2005reinforcement}
Peter Stone, Richard~S Sutton, and Gregory Kuhlmann.
\newblock Reinforcement learning for robocup soccer keepaway.
\newblock \emph{Adaptive Behavior}, 13\penalty0 (3):\penalty0 165--188, 2005.

\bibitem[Taylor and Stone(2009)]{taylor2009transfer}
Matthew~E Taylor and Peter Stone.
\newblock Transfer learning for reinforcement learning domains: A survey.
\newblock \emph{Journal of Machine Learning Research}, 10\penalty0 (7), 2009.

\bibitem[Thomas et~al.(2017)Thomas, Pondard, Bengio, Sarfati, Beaudoin, Meurs, Pineau, Precup, and Bengio]{thomas2017independently}
Valentin Thomas, Jules Pondard, Emmanuel Bengio, Marc Sarfati, Philippe Beaudoin, Marie-Jean Meurs, Joelle Pineau, Doina Precup, and Yoshua Bengio.
\newblock Independently controllable factors.
\newblock \emph{arXiv preprint arXiv:1708.01289}, 2017.

\bibitem[Tsai et~al.(2018)Tsai, Liang, Zadeh, Morency, and Salakhutdinov]{tsai2018learning}
Yao-Hung~Hubert Tsai, Paul~Pu Liang, Amir Zadeh, Louis-Philippe Morency, and Ruslan Salakhutdinov.
\newblock Learning factorized multimodal representations.
\newblock \emph{arXiv preprint arXiv:1806.06176}, 2018.

\bibitem[Wei et~al.(2017)Wei, Wang, and Zhu]{wei2017deep}
Tianshu Wei, Yanzhi Wang, and Qi~Zhu.
\newblock Deep reinforcement learning for building hvac control.
\newblock In \emph{Proceedings of the 54th annual design automation conference 2017}, pages 1--6, 2017.

\bibitem[Xian et~al.(2018)Xian, Lampert, Schiele, and Akata]{xian2018zero}
Yongqin Xian, Christoph~H Lampert, Bernt Schiele, and Zeynep Akata.
\newblock Zero-shot learning—a comprehensive evaluation of the good, the bad and the ugly.
\newblock \emph{IEEE transactions on pattern analysis and machine intelligence}, 41\penalty0 (9):\penalty0 2251--2265, 2018.

\bibitem[Zhang et~al.(2018)Zhang, Vinyals, Munos, and Bengio]{zhang2018study}
Chiyuan Zhang, Oriol Vinyals, Remi Munos, and Samy Bengio.
\newblock A study on overfitting in deep reinforcement learning.
\newblock \emph{arXiv preprint arXiv:1804.06893}, 2018.

\end{thebibliography}

\clearpage
\appendix

\section{Additional Figures}
\label{appendix:a}

\begin{figure}[h] 
    \centering 
\begin{tikzpicture}[>=Stealth, node distance=1cm]
    
    \node[draw, ellipse, minimum width=1.2cm, minimum height=0.7cm, thick, align=center] (Ct) {distance\\matrix, $t$};
    \node[draw, ellipse, below=of Ct, minimum width=1.2cm, minimum height=0.7cm, thick, align=center] (X4t) {grid size, $t$};
    \node[draw, ellipse, below=of X4t, minimum width=1.2cm, minimum height=0.7cm, thick, align=center] (X1t) {goal\\location, $t$};
    \node[draw, ellipse, below=of X1t, minimum width=1.2cm, minimum height=0.7cm, thick, align=center] (X2t) {hole\\locations, $t$};
    \node[draw, ellipse, below=of X2t, minimum width=1.2cm, minimum height=0.7cm, thick, align=center] (X3t) {agent\\location, $t$};

    \node[draw, ellipse, right=of Ct, minimum width=1.2cm, minimum height=0.7cm, thick, align=center] (Ctp1) {distance\\matrix, $t+1$};
    \node[draw, ellipse, below=of Ctp1, minimum width=1.2cm, minimum height=0.7cm, thick, align=center] (X4tp1) {grid size, $t+1$};
    \node[draw, ellipse, below=of X4tp1, minimum width=1.2cm, minimum height=0.7cm, thick, align=center] (X1tp1) {goal\\location, $t+1$};
    \node[draw, ellipse, below=of X1tp1, minimum width=1.2cm, minimum height=0.7cm, thick, align=center] (X2tp1) {hole\\locations, $t+1$};
    \node[draw, ellipse, below=of X2tp1, minimum width=1.2cm, minimum height=0.7cm, thick, align=center] (X3tp1) {agent\\location, $t+1$};
    
    \draw[->, black] (X1t) -- (X1tp1);
    \draw[->, black] (X2t) -- (X2tp1);
    \draw[->, black] (X3t) -- (X3tp1);
    \draw[->, black] (X4t) -- (X4tp1);
    
    \draw[->, black] (X4t) -- (X3tp1);
    \draw[->, black] (X2t) -- (X3tp1);
    \draw[->, black] (Ct) -- (Ctp1);
    \end{tikzpicture}
\caption{A Dynamic Bayesian Network for the factored MDP of the Shifting Frozen Lake. The distance matrix (from the goal location), the grid size, the goal location and the hole locations are constant throughout each episode.}
  \label{fig:sfl-dbn}
\end{figure}

\end{document}